# Prior-Guided Frequency-Calibrated Virtual EEG Channel Inference from Four Frontal Electrodes for Wearable EEG Augmentation

Minghao Xiao

School of Biomedical Science and Engineering, South China University of Technology, Guangzhou 511442, China

School of Automation Science and Engineering, South China University of Technology, Guangzhou 510640, China

## Abstract

Low-channel wearable electroencephalography (EEG) is attractive for long-term monitoring, but four frontal electrodes provide only a sparse and spatially biased sampling of the scalp potential field. Virtual-channel methods should therefore be framed not as recovery of independent unmeasured brain activity, but as prior-guided conditional inference of posterior predictive scalp-potential representations at target electrode locations. We present FAVC-Net, a compact frequency-calibrated virtual-channel inference network that estimates 13 target channels from Fp1, Fp2, F7, and F8. The model combines shared multi-scale source encoding, source-state embeddings, target-conditioned signed source-block mixing, GATv2-based attention refinement, attention-consistent skip fusion, and weak Welch power spectral density calibration. The generator is trained as a task-agnostic reconstruction module, without class-label, classification, or CSP-like discriminative constraints, so that the virtual montage remains tied to conditional scalp-potential estimation rather than to a specific downstream decision. On the PRED+CT dataset, FAVC-Net achieved the best joint waveform-spectral operating point among neural and interpolation baselines. Its time-domain gains were modest, whereas log-spectral distance and PSD KL divergence were reduced by 30.50% and 38.94% relative to the strongest non-FAVC comparator. Under wearable-like source perturbations, the model preserved spectral fidelity and channel-frequency texture, with anti-collapse benefits most evident under EMG-like bursts and mixed stress. These results support virtual EEG channels as montage-compatible, frequency-calibrated posterior predictive representations derived from sparse frontal measurements, not as independent substitutes for physically recorded electrodes.

## 1. Introduction

Electroencephalography (EEG) provides millisecond-scale access to neural dynamics and remains important in neuroengineering, affective computing, brain-computer interfaces, and quantitative neurophysiology [6, 24]. Dense or clinical-grade montages provide broad scalp coverage, but their setup time, electrode maintenance, cost, and wearing burden limit ambulatory and long-term wearable use. Four-frontal systems are therefore attractive for practical monitoring, yet they observe only a partial and spatially biased sampling of the scalp potential field. This limitation matters because many EEG signatures used in practical analysis are expressed through distributed temporal, spectral, and topographic patterns rather than through isolated frontal amplitudes alone [1, 3, 11, 21].

A careful problem definition is essential. A scalp electrode is not a direct measurement of a single cortical region; it records a reference-dependent mixture of neural and non-neural electrical sources shaped by volume conduction, head geometry, electrode placement, and artifacts [6, 24]. Estimating an unobserved electrode from four frontal electrodes is therefore not equivalent to generating posterior or parietal brain activity from frontal brain activity. It is a conditional spatial-completion problem on a sampled scalp potential field: the output should be interpreted as the model's best target-electrode representation consistent with the measured frontal segment and with priors learned from multichannel EEG data.

This framing also sets an information-theoretic boundary. Once a trained deterministic generator is fixed, its virtual channels are functions of the measured source channels and model parameters; they cannot create independent measurement information that was absent from the current low-channel recording. Their value comes from prior transfer and representation reshaping: the training dataset can encode spatial topology, cross-channel statistics, spectral regularities, reference-dependent coupling, and population-level structure, and the deployed model can use those priors to map sparse EEG into a montage-compatible representation. The relevant scientific question is therefore not whether a virtual channel is identical to a physical channel, but which target channels, frequency ranges, and feature types remain conditionally inferable from the available source montage.

Classical interpolation methods such as nearest-neighbor interpolation, inverse-distance weighting, and spherical splines are transparent and useful when a dense montage contains only a small number of defective electrodes [8, 9, 12, 13, 22]. Four frontal electrodes, however, do not define a conventional bad-channel interpolation setting. They define a strongly underdetermined conditional reconstruction problem in which 13 target-electrode representations must be inferred from a spatially clustered sparse montage. In this regime, the mapping depends on source state, reference effects, oscillatory regime, subject variability, and nonstationary source-target coupling, not only on geometric distance.

Recent learned approaches have explored EEG spatial super-resolution, channel restoration, attention-based interpolation, state-space models, Transformer architectures, diffusion models, and attention U-Nets [5, 7, 16-19, 26, 27, 30-33]. These studies establish the feasibility of learned dense-channel estimation, but they also reveal an evaluation gap. Many comparisons emphasize waveform reconstruction scores, whereas wearable EEG analysis often relies on PSD shape, band powers, spectral slope, regional energy gradients, and channel-frequency texture [10, 15, 23, 29]. A virtual channel can achieve acceptable amplitude error or Pearson correlation while misallocating spectral energy or producing overly homogeneous target channels. Sparse-to-dense EEG inference should therefore be treated as a dual-domain reconstruction task rather than as pure waveform matching.

We propose FAVC-Net, a frequency-calibrated virtual-channel inference network that estimates 13 target EEG channels from Fp1, Fp2, F7, and F8. The design follows three principles. First, source-to-target mapping should be state-dependent because the same frontal electrodes can reflect different oscillatory regimes across windows. Second, target estimation should be channel-specific because frontal, central, temporal, and posterior targets require different source-response rules. Third, source mixing should allow signed contributions because EEG referencing and phase relationships can induce negative dependencies that are suppressed by positive-simplex attention.

The contributions are threefold: (i) a compact target-conditioned architecture for prior-guided four-frontal-to-multichannel EEG representation completion; (ii) a dual-domain training and evaluation strategy that treats waveform fidelity and spectral-topographic fidelity as joint criteria while excluding task-label supervision from generator training; and (iii) a checkpoint-only corrupted-source robustness analysis showing that the model preserves absolute spectral fidelity and resists spectral collapse under wearable-relevant perturbations. The claim is intentionally bounded: the estimated channels are deterministic, frequency-calibrated posterior predictive representations conditioned on sparse frontal measurements and learned multichannel priors, not independent substitutes for physically recorded posterior electrodes.

## 2. Methods

### 2.1 Data, channel selection, and preprocessing

The primary reconstruction experiment used the PRED+CT EEG database released by Cavanagh et al. [4]. The database contains recordings from 119 subjects performing a probabilistic learning task. According to the dataset description used in this analysis, participants comprised 74 male and 45 female subjects aged 18-24 years; EEG was recorded at 500 Hz from 66 electrodes arranged according to the international 10-20 system [13]. Trait-anxiety categories based on the Spielberger Trait Anxiety Inventory were recorded in the database [25], but these labels were not used for reconstruction training, validation, testing, model selection, robustness analysis, or virtual-channel generation.

Seventeen channels were selected to cover frontal, temporal, central, and parietal regions: Fp1, Fp2, F7, F3, Fz, F4, F8, T3, C3, Cz, C4, T4, T5, P3, Pz, P4, and T6. The measured source set was Fp1, Fp2, F7, and F8, representing a practical four-frontal wearable layout. The target set contained the remaining 13 channels. Signals were band-pass filtered from 0.5 to 45 Hz and segmented into 3000-sample windows, corresponding to 6 s at 500 Hz. The reconstruction split was strictly subject-independent: 95 subjects and 950 segments were used for training, 11 subjects and 110 segments for validation, and 13 subjects and 130 segments for testing. Channel-wise standard deviations were computed from the training subjects only and were used for normalization and normalized MAE reporting. Acquisition, montage, filtering, segmentation, and split information are reported in line with general EEG/MEG publication recommendations [14].

### 2.2 Prior-guided problem formulation

Let S denote the measured source set, ordered as Fp1, Fp2, F7, and F8, and let T denote the 13 target-electrode set. For each 6-s segment, the sparse source observation and target montage are

$$X = [x_1, x_2, x_3, x_4]^\top \in \mathbb{R}^{4\times T}, \qquad Y = [y_1, \dots, y_{13}]^\top \in \mathbb{R}^{13\times T}.$$

FAVC-Net implements a deterministic point estimator of the conditional target representation from the measured sources and target identity:

$$\hat{Y} = G_\theta(X, q) = [\hat{y}_1, \dots, \hat{y}_{13}]^\top \in \mathbb{R}^{13\times T},$$

where the target-conditioning term denotes the target identity or the target-specific parameters used by the generator. More generally, the desired object is the conditional posterior predictive distribution of unobserved target-electrode signals given the measured source segment, the montage geometry, and the multichannel EEG prior encoded in the trained model. The deterministic output used here is a point estimate of that distribution, not an independent physical recording.

At inference, target-channel waveforms, target-channel statistics, and task labels are unavailable. The model must therefore infer target-specific source mixtures from measured source EEG and electrode identity alone. Because the estimated montage is a function of the measured source segment and learned parameters, it should not be described as creating new measurement information. It is better interpreted as a high-density posterior predictive representation that transfers training-set priors into the low-channel setting.

The reconstruction task is deliberately separated from downstream task discrimination. No classification labels, task losses, CSP-like discriminative losses, or label-conditioned priors are used for generator training, checkpoint selection, robustness analysis, or target-channel estimation. This restriction keeps the virtual channels aligned with conditional scalp-potential estimation and reduces the risk that the generated montage encodes task-specific decision features rather than physiologically interpretable source-conditioned structure.

## 2.3 FAVC-Net architecture

FAVC-Net is shown in Fig. 1. Each measured source channel is passed through the same one-dimensional multi-scale encoder. The encoder has four stride-2 stages and parallel temporal kernels of length 3, 5, and 9, with output widths 32, 64, 128, and 256. Weight sharing keeps the model compact and prevents the four measured sources from being represented by unrelated feature extractors, while the multi-scale kernels expose local transients, rhythmic envelopes, alpha/beta oscillations, and nonstationary artifacts at different receptive-field widths.

For source $i$, the shared encoder produces

$$h_i = E_\phi(x_i) \in \mathbb{R}^{C \times L}, \qquad i = 1, \dots, 4.$$

The final $C = 256$ channels are divided into $B = 32$ latent blocks. This block partition lets a virtual target select source contributions at a frequency- and morphology-sensitive latent-block level instead of relying on a single scalar source weight.

The current source state is summarized by temporal moments and extrema of the encoded feature map:

$$s_i = \left[\mu_\tau(h_i),\ \sigma_\tau(h_i),\ \max_\tau h_i,\ \min_\tau h_i\right].$$

A compact identity network maps this descriptor to a 32-dimensional source-state embedding:

$$e_i = \mathrm{LN}\big(W_2\, \delta\big(\mathrm{BN}(W_1 s_i + b_1)\big) + b_2\big) \in \mathbb{R}^{32}, \qquad E_s = [e_1; e_2; e_3; e_4].$$

Here $\delta(\cdot)$ is a pointwise nonlinearity, BN denotes batch normalization, and LN denotes layer normalization. The embeddings summarize the state of the measured sparse montage and drive attention generation rather than using target waveforms, which are unavailable at inference.

For virtual target $t$, a target-private first layer followed by a shared second layer produces source-by-block attention scores:

$$A_t = \mathrm{reshape}_{4 \times B}\left(W_a^{(2)}\, \delta\left(W_{a,t}^{(1)} E_s + b_{a,t}^{(1)}\right) + b_a^{(2)}\right), \qquad A_t \in \mathbb{R}^{4 \times B}.$$

The private first layer gives each virtual electrode a distinct source-response rule, while the shared second layer encourages the latent blocks to retain comparable meaning across targets. This balances specialization and statistical efficiency: Fz, T3, Cz, Pz, and T6 should not share one global source mixture, but all targets are still inferred from the same four measured sources. The target-private projection is also the main channel-specific parameterization in the generator; the source encoder, shared attention layer, graph refinement, adapter, and decoder are reused across targets.

The 13 virtual targets are then treated as graph nodes. For GATv2 refinement, each source-by-block attention tensor is flattened into a node feature. This refinement follows graph-attention principles [28] and uses the more expressive GATv2 formulation [2]:

$$z_t = \mathrm{flatten}(A_t) \in \mathbb{R}^{4B}.$$

Approximate 10-20 coordinates define a weak spatial prior over virtual targets:

$$P_{tu} = \exp\left(-\frac{\| p_t - p_u \|_2^2}{\tau_p}\right),$$

where $p_t$ and $p_u$ are two-dimensional electrode coordinates and $\tau_p$ controls the spatial scale. The prior is used only as a neighborhood mask and a small logit bias; it is not an output-voltage smoothing constraint. For connected nodes $u \in \mathcal{N}(t)$, the GATv2 refinement logit is

$$r_{tu} = a_g^\top \mathrm{LeakyReLU}\big(W_s z_t + W_u z_u + b_g\big) + \lambda_g \log(P_{tu} + \epsilon),$$

and the attention coefficient is

$$\alpha_{tu} = \frac{\exp(r_{tu})}{\sum_{v\in\mathcal{N}(t)} \exp(r_{tv})}.$$

The graph message and gated residual update are

$$m_t = \sum_{u\in\mathcal{N}(t)} \alpha_{tu} W_m z_u, \qquad \gamma_t = \sigma\big(W_\gamma [z_t \parallel m_t] + b_\gamma\big), \qquad z_t^{\mathrm{ref}} = z_t + \gamma_t \odot m_t.$$

The refined vector is reshaped back to a source-by-block tensor before source aggregation. Thus, GATv2 exchanges information about plausible source-block patterns among virtual targets, but it does not average reconstructed voltages across neighboring electrodes. This distinction is important because direct smoothing may improve visually pleasing topographies while degrading channel-specific spectral texture.

After graph refinement, scores are normalized over the four source channels by signed $L_1$ normalization rather than softmax:

$$\tilde{a}_{tib} = \frac{a_{tib}^{\mathrm{ref}}}{\sum_{j=1}^{4} \left|a_{tjb}^{\mathrm{ref}}\right| + \epsilon}, \qquad i = 1, \dots, 4,\ \ b = 1, \dots, B.$$

Let $b(k)$ map encoder channel $k$ to its latent block. The target-specific latent state is

$$\tilde{h}_{t,k,:} = \sum_{i=1}^{4} \tilde{a}_{ti,b(k)}\, h_{i,k,:}.$$

Signed normalization allows negative or phase-inverted source-target relationships under a fixed EEG reference, which would be suppressed by positive-simplex softmax attention. A shared $1 \times 1$ convolutional adapter and shared transposed-convolution decoder estimate the 13 virtual waveforms. Shallow encoder features are compressed according to learned attention magnitude and fused into decoder stages as attention-consistent skip connections. The final implementation contains 912,770 trainable parameters.

This factorization gives the backbone a useful modular structure. Most parameters are shared across target channels, whereas target identity mainly enters through a compact target-specific attention generator and, for graph refinement, through target-coordinate neighborhood information. Consequently, the architecture is conceptually extensible to target montages with different channel sets by introducing or adapting target-specific attention parameters and spatial metadata while reusing the shared encoder-decoder backbone. Such extensions would still require validation, and usually training or fine-tuning, on data containing the additional target locations; the present experiments substantiate the claim only for the 13 target channels studied here.

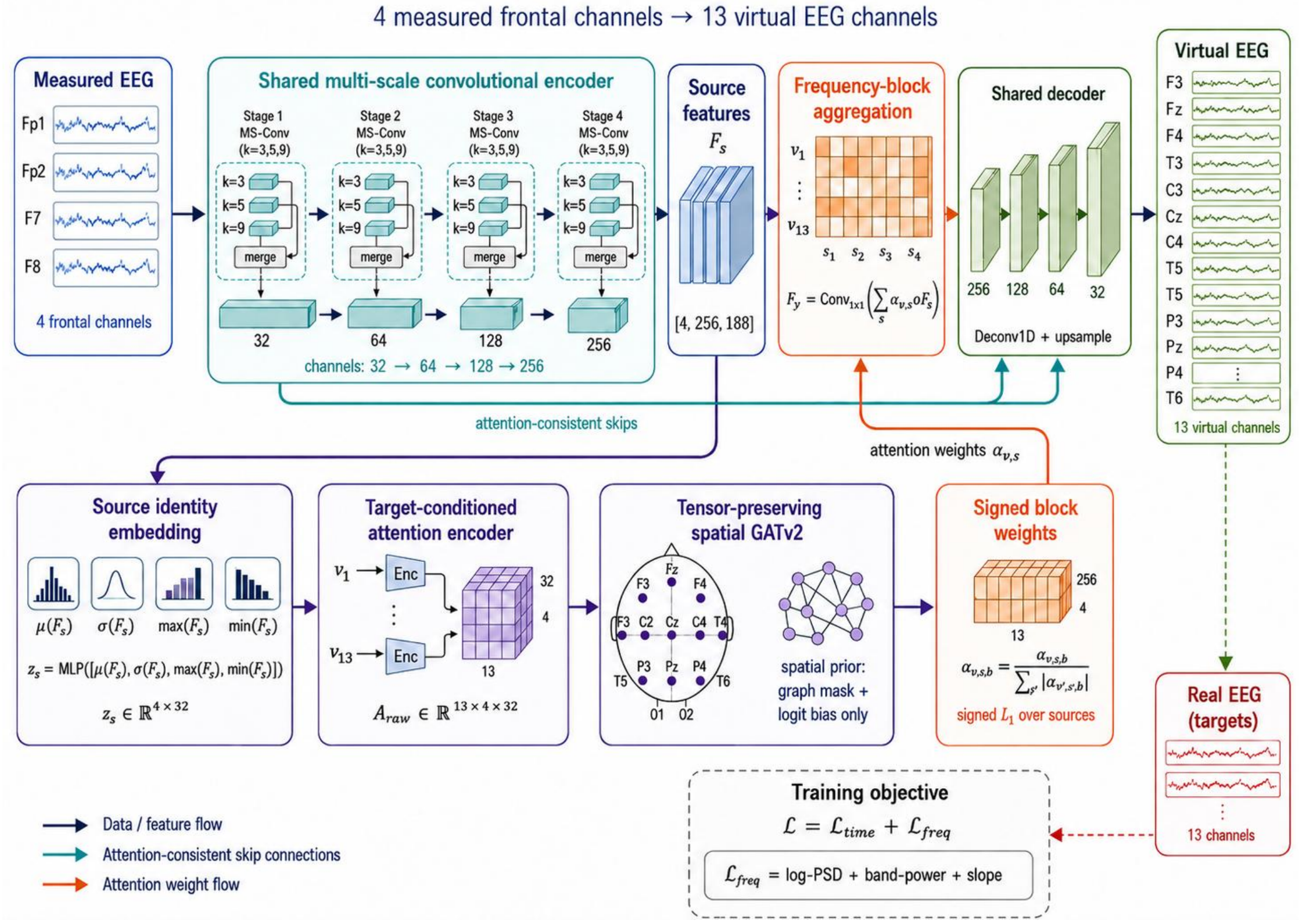

**Fig. 1 FAVC-Net architecture. Four frontal source channels are encoded by a shared multi-scale encoder. Source-state embeddings drive target-conditioned signed attention. The source-by-block attention tensor is refined over virtual targets by GATv2 and then used for signed aggregation, attention-consistent skip fusion, and shared decoding into 13 virtual EEG channel estimates.**

## 2.4 Training objective and reconstruction metrics

The training objective is reconstruction-based and task-agnostic. It combines a dominant normalized waveform term with weak frequency-domain calibration, and it does not include classification losses, CSP-like discriminative terms, or other task-prior constraints:

$$\mathcal{L} = 0.90\,\mathcal{L}_{\text{wave}} + 0.10\,\mathcal{L}_{\text{PSD}}.$$

The waveform term is a normalized absolute reconstruction loss:

$$\mathcal{L}_{\text{wave}} = \frac{1}{13T}\sum_{c=1}^{13}\sum_{t=1}^{T}\frac{|\hat{y}_{c,t} - y_{c,t}|}{\sigma_c + \epsilon},$$

where $\sigma_c$ is the training-set standard deviation of target channel $c$. The PSD term is computed from differentiable Welch estimates over 0.5-45 Hz. For a Hann window $w$, sampling frequency $f_s$, and windowed frame $x_{c,n}$, the PSD estimate is

$$S_c(f) = \frac{1}{N}\sum_{n=1}^{N}\frac{|\mathcal{F}\{w \odot x_{c,n}\}(f)|^2}{f_s \sum_\ell w_\ell^2}.$$

The implemented PSD calibration combines log-PSD shape, relative physiological band allocation, and spectral-slope consistency:

$$\mathcal{L}_{\text{PSD}} = \lambda_{\log} \parallel \log(\hat{S} + \epsilon) - \log(S + \epsilon) \parallel_1 + \lambda_{\text{band}} \parallel \hat{\rho} - \rho \parallel_1 + \lambda_{\text{slope}} \parallel \hat{\beta} - \beta \parallel_1.$$

The calibration term is intentionally weak: it is designed to correct spectral allocation without allowing PSD matching to dominate sample-wise waveform reconstruction. It is therefore a measurement-domain spectral constraint rather than a downstream task prior. A validation sweep showed a smooth trade-off in which increasing the PSD weight gradually improved spectral metrics while slightly reducing waveform fidelity. The 0.90/0.10 setting was selected as a balanced operating point rather than as a narrowly tuned optimum.

Four reconstruction metrics were used. Normalized MAE measures amplitude error relative to training-set channel standard deviation:

$$\mathrm{nMAE} = \frac{1}{13T}\sum_{c=1}^{13}\sum_{t=1}^{T}\frac{\left|\hat{y}_{c,t} - y_{c,t}\right|}{\sigma_c + \epsilon}.$$

Pearson correlation measures phase-sensitive waveform agreement:

$$r = \frac{1}{13}\sum_{c=1}^{13}\frac{\sum_t \left(\hat{y}_{c,t} - \bar{\hat{y}}_c\right)\left(y_{c,t} - \bar{y}_c\right)}{\sqrt{\sum_t \left(\hat{y}_{c,t} - \bar{\hat{y}}_c\right)^2 \sum_t \left(y_{c,t} - \bar{y}_c\right)^2} + \epsilon}.$$

Log-spectral distance compares the absolute log-power envelope:

$$\mathrm{LSD} = \sqrt{\frac{1}{13|\mathcal{F}|}\sum_{c=1}^{13}\sum_{f\in\mathcal{F}}\left[\log\left(\hat{S}_c(f) + \epsilon\right) - \log\left(S_c(f) + \epsilon\right)\right]^2}.$$

PSD-normalized KL divergence compares within-channel frequency allocation:

$$p_c(f) = \frac{S_c(f)}{\sum_{f'\in\mathcal{F}} S_c(f') + \epsilon}, \qquad \hat{p}_c(f) = \frac{\hat{S}_c(f)}{\sum_{f'\in\mathcal{F}} \hat{S}_c(f') + \epsilon},$$

$$\mathrm{KL} = \frac{1}{13}\sum_{c=1}^{13}\sum_{f\in\mathcal{F}} p_c(f)\log\frac{p_c(f) + \epsilon}{\hat{p}_c(f) + \epsilon}.$$

Together, these metrics separate amplitude fidelity, phase-sensitive waveform agreement, absolute log-power structure, and normalized spectral-shape similarity. Raw-scale MAE was derived from the same training-set channel standard deviations for all methods, so differences in raw MAE are not caused by method-specific rescaling.

## 2.5 Baselines and implementation protocol

FAVC-Net was compared with neural and interpolation baselines: Deep-CNN, GN, EC-Informer, VEEG-A-U-Net, ESTformer, inverse-distance weighting, spherical-spline interpolation, and nearest-neighbor interpolation. All methods used the same source channels, target channels, subject-level split, and evaluation metrics. Neural methods were summarized as mean ± standard deviation across repeated runs. FAVC-Net was optimized with AdamW [20], learning rate 1.0 × 10^-4, weight decay 1.0 × 10^-5, batch size 16, gradient clipping at 1.0, ReduceLROnPlateau scheduling, and early stopping.

## 2.6 Checkpoint-only robustness to wearable-like source-channel perturbations

To test whether virtual-channel estimation remains useful when measured frontal inputs are corrupted, robustness analysis was performed in a checkpoint-only manner. Saved checkpoints were evaluated without retraining, fine-tuning, noise-specific calibration, or post hoc checkpoint selection. Perturbations were applied only to the four measured source channels, whereas the 13 target channels remained clean. The experiment therefore measured the ability to reconstruct the clean target montage from imperfect wearable inputs, not the ability to copy noise into the targets.

The main text focuses on four perturbations that are practically relevant after routine preprocessing: moderate EMG-like high-frequency bursts, 0.5-s source-channel dropout, ±20% channel-wise gain mismatch, and moderate mixed wearable stress. The moderate mixed condition applied AWGN at 10 dB, EMG bursts at 10 dB with two bursts of 0.30-0.80 s and channel probability 0.50, one 0.50-s source-channel dropout, and ±20% gain mismatch. Absolute noisy-condition performance was used as the primary robustness endpoint because percentage degradation can over-penalize a model that starts from a substantially lower clean spectral error.

The perturbations were generated reproducibly from the split seed, condition name, and repeat index. AWGN was scaled per sample and source channel according to source RMS and the specified SNR:

$$\sigma_{b,c} = \frac{\mathrm{RMS}\left(x_{b,c,:}\right)}{\sqrt{10^{\mathrm{SNR}_{\mathrm{dB}}/10}}}, \qquad x'_{b,c,t} = x_{b,c,t} + \varepsilon_{b,c,t}, \qquad \varepsilon_{b,c,t} \sim \mathcal{N}\left(0, \sigma_{b,c}^{2}\right).$$

EMG-like bursts were generated as 20-45 Hz band-limited noise multiplied by a Hann temporal envelope and RMS-calibrated to the target SNR. Dropout selected one source channel and set a contiguous temporal window to zero,

$$x'_{b,c,t} = m_{b,c,t} x_{b,c,t}, \qquad m_{b,c,t} \in \{0,1\},$$

and gain mismatch multiplied each source by a random channel-wise factor:

$$g_{b,c} \sim \mathrm{Uniform}(1-\rho, 1+\rho), \qquad x'_{b,c,t} = g_{b,c} x_{b,c,t}.$$

Robustness was quantified by LSD, PSD KL divergence, spectral-collapse index (SCI), and channel-frequency texture correlation (CFTC). CFTC is the Pearson correlation between vectorized log-PSD tensors of predicted and true target channels:

$$\mathrm{CFTC}_{b} = \mathrm{corr}\left(\mathrm{flatten}\{\log\left(\hat{S}_{b} + \epsilon\right)\}, \mathrm{flatten}\{\log\left(S_{b} + \epsilon\right)\}\right).$$

SCI measures whether predicted channels become spectrally over-homogeneous. For each subject-window $b$ and all target-channel pairs $i < j$, the pairwise channel spectral distances are

$$D_{b}^{\mathrm{pred}} = \frac{2}{C(C-1)} \sum_{i<j} \left\| \log\left(\hat{S}_{b,i,:} + \epsilon\right) - \log\left(\hat{S}_{b,j,:} + \epsilon\right) \right\|_{2},$$

$$D_{b}^{\mathrm{true}} = \frac{2}{C(C-1)} \sum_{i<j} \left\| \log\left(S_{b,i,:} + \epsilon\right) - \log\left(S_{b,j,:} + \epsilon\right) \right\|_{2}, \qquad C = 13.$$

The pairwise collapse component penalizes loss of between-channel spectral diversity:

$$\mathrm{SCI}_{b}^{\mathrm{pair}} = \max\left(1 - \frac{D_{b}^{\mathrm{pred}}}{D_{b}^{\mathrm{true}} + \epsilon},\ 0\right).$$

For band-topographic variance, let $\mathcal{B}_{k}$ denote the delta, theta, alpha, beta, or low-gamma band. Band powers and variance ratios are

$$\mathrm{BP}_{b,c,k}^{\mathrm{pred}} = \log\left(\sum_{f \in \mathcal{B}_{k}} \hat{S}_{b,c}(f) + \epsilon\right), \qquad \mathrm{BP}_{b,c,k}^{\mathrm{true}} = \log\left(\sum_{f \in \mathcal{B}_{k}} S_{b,c}(f) + \epsilon\right),$$

$$\mathrm{BTVR}_{b,k} = \frac{\mathrm{std}_{c}\left(\mathrm{BP}_{b,c,k}^{\mathrm{pred}}\right)}{\mathrm{std}_{c}\left(\mathrm{BP}_{b,c,k}^{\mathrm{true}}\right) + \epsilon}.$$

The topographic-collapse component and implemented SCI are

$$\mathrm{SCI}_b^{\mathrm{topo}} = \frac{1}{K}\sum_{k=1}^{K} \max\left(1 - \mathrm{BTVR}_{b,k}, 0\right), \qquad \mathrm{SCI}_b = 0.5\,\mathrm{SCI}_b^{\mathrm{pair}} + 0.5\,\mathrm{SCI}_b^{\mathrm{topo}}.$$

Lower SCI indicates less spectral collapse. Robustness-condition summaries were expressed as mean ± 95% confidence intervals over the aggregated evaluation units, and percentage advantages were computed against the strongest non-FAVC comparator for each endpoint under each perturbation.

### 2.7 Exploratory external representation transfer check

An auxiliary downstream experiment examined whether the fixed virtual montage provides useful representation value outside the reconstruction dataset. The external recordings came from the low-cost, wireless, four-channel EEG measurement system for virtual-reality environments described by Yu and Guo [34]. Thirty-four subjects were retained after exclusion, each contributing 30 six-second windows. The device recorded four frontal channels at 250 Hz; AF7 and AF8 were used as lateral-frontal proxies for F7 and F8. External signals were filtered using the same passband logic, resampled to the 500 Hz/3000-sample model input length, and normalized with the PRED+CT training-set channel statistics stored in the reconstruction checkpoint. The generator was used zero-shot and was never updated with external data.

The external dataset contained only the four measured frontal channels and therefore had no ground-truth recordings for the 13 target channels. Consequently, no external target-channel waveform, target-channel statistic, or target-derived preprocessing parameter was available to the generator or to the downstream pipeline. Five channel configurations were evaluated: D0, real four-channel features only; D1, real channels plus 13 FAVC-estimated virtual-channel features; D2, estimated virtual channels only; D3, real channels plus duplicated/proxy-expanded channels; and D4, real channels plus phase-random pseudo-channels. D3 and D4 served as negative controls for dimensionality expansion and arbitrary pseudo-channel spectra. This experiment is reported as exploratory representation-transfer evidence rather than clinical validation.

Feature extraction, feature screening, dimensionality reduction, and classifier training were performed strictly within each subject-held-out cross-validation fold. The held-out subject was excluded from ANOVA screening, PCA fitting, classifier training, and all model-selection operations in that fold. Labels were used only within the downstream training portion of each fold and were never used to train, tune, or select the virtual-channel generator. This design prevents target-channel leakage, subject leakage, and test-fold leakage while testing whether the fixed posterior predictive montage can improve finite-sample downstream representations under device, montage, sampling-rate, and cohort mismatch.

## 3. Results

### 3.1 Overall dual-domain reconstruction performance

FAVC-Net achieved the best joint waveform-spectral operating point among the evaluated methods (Table 1). Relative to the strongest non-FAVC comparator for each metric, it reduced nMAE from 0.1497 to 0.1463, reduced raw MAE from 4.655 μV to 4.548 μV, and improved Pearson correlation from 0.6206 to 0.6245. These time-domain gains are deliberately interpreted as modest. The main advantage was spectral: LSD decreased from 0.9657 to 0.6711, and PSD KL divergence decreased from 0.1506 to 0.0920, corresponding to 30.50% and 38.94% relative reductions. Thus, the method's contribution is not simply a lower pointwise error but a more faithful spectral-topographic posterior predictive representation under a comparable waveform-fidelity constraint.

The result is also parameter efficient. FAVC-Net used 0.913 million parameters, far fewer than GN, VEEG-A-U-Net, and Deep-CNN, while outperforming them in the combined metric set. The interpolation baselines remained useful references but were disadvantaged by the sparse and spatially clustered source montage. Their errors show that the task is not equivalent to replacing one missing electrode in a dense array; frontal-to-posterior inference requires learned conditional structure beyond smooth geometric interpolation.

**Table 1 Compact overall reconstruction performance on 13 virtual target channels. Values are mean ± standard deviation; lower values are better except Pearson r.**

| Method | Parameters | nMAE ↓ | Raw MAE (µV) ↓ | Pearson r ↑ | LSD ↓ | PSD KL ↓ |
|---|---|---|---|---|---|---|
| FAVC-Net | 912,770 | 0.1463 ± 0.0033 | 4.548 ± 0.102 | 0.6245 ± 0.0082 | 0.6711 ± 0.0076 | 0.0920 ± 0.0030 |
| GN | 4,913,473 | 0.1497 ± 0.0045 | 4.655 ± 0.138 | 0.6206 ± 0.0114 | 1.5574 ± 0.1146 | 0.1876 ± 0.0081 |
| ESTformer | 874,162 | 0.1536 ± 0.0032 | 4.777 ± 0.099 | 0.5804 ± 0.0119 | 1.7495 ± 0.0725 | 0.1883 ± 0.0025 |
| VEEG-A-U-Net | 5,940,161 | 0.1607 ± 0.0039 | 4.992 ± 0.122 | 0.5944 ± 0.0127 | 0.9657 ± 0.0468 | 0.1652 ± 0.0219 |
| Deep-CNN | 2,255,181 | 0.1620 ± 0.0059 | 5.038 ± 0.183 | 0.5924 ± 0.0096 | 1.2961 ± 0.1639 | 0.1506 ± 0.0120 |
| EC-Informer | 600,304 | 0.1651 ± 0.0062 | 5.141 ± 0.197 | 0.5321 ± 0.0340 | 2.1307 ± 0.1780 | 0.2846 ± 0.0351 |
| IDW | 0 | 0.3233 ± 0.0131 | 10.008 ± 0.411 | 0.3681 ± 0.0269 | 1.3078 ± 0.0691 | 0.2897 ± 0.0203 |
| NNI | 0 | 0.3341 ± 0.0143 | 10.350 ± 0.449 | 0.4126 ± 0.0194 | 1.4965 ± 0.0350 | 0.2462 ± 0.0231 |
| SSI | 0 | 0.5115 ± 0.0346 | 15.761 ± 1.068 | 0.4999 ± 0.0212 | 2.3992 ± 0.1318 | 0.2458 ± 0.0287 |

nMAE, normalized mean absolute error; MAE, mean absolute error; LSD, log-spectral distance; KL, PSD-distribution KL divergence.

## 3.2 Channel-wise observability and waveform visualization

Channel-wise results (Table 2) show a plausible observability gradient. Near-frontal targets F3, Fz, and F4 had the highest correlations because they are closest to the measured frontal sources and share frontal activity. Lateral temporal and central channels remained moderately recoverable. Posterior and parietal targets had lower phase-sensitive correlations, which is expected from four frontal sources. The representative overlay in Fig. 2 visually confirms that estimated signals follow large-scale morphology and local waveform patterns in many channels, while posterior targets retain plausible but less phase-locked variability.

Raw MAE values varied across targets, reflecting both reconstruction difficulty and intrinsic channel amplitude scale. The high correlations for F3, Fz, and F4 should not be generalized to all targets, and the lower posterior correlations should not be interpreted as model collapse. They instead define the practical observability boundary of a four-frontal sensor layout. This boundary is useful for deployment because it indicates which virtual channels are most reliable for phase-sensitive analyses and which should be interpreted primarily as spectral-topographic representations.

**Table 2 Channel-wise FAVC-Net reconstruction results under the 0.90/0.10 objective.**

| Channel | nMAE ↓ | Raw MAE (µV) | Pearson r ↑ | Observability interpretation |
|---|---|---|---|---|
| F3 | 0.1298 ± 0.0028 | 4.170 ± 0.089 | 0.8820 ± 0.0037 | Near-frontal; strong source observability |
| Fz | 0.1205 ± 0.0015 | 3.870 ± 0.050 | 0.9025 ± 0.0018 | Near-frontal; strong source observability |
| F4 | 0.1188 ± 0.0022 | 3.936 ± 0.073 | 0.9083 ± 0.0033 | Near-frontal; strong source observability |
| T3 | 0.1775 ± 0.0033 | 5.549 ± 0.102 | 0.6796 ± 0.0062 | Lateral temporal; moderate frontal observability |
| C3 | 0.1085 ± 0.0020 | 3.228 ± 0.059 | 0.5779 ± 0.0084 | Central; moderate or low-amplitude observability |
| Cz | 0.0576 ± 0.0007 | 1.714 ± 0.022 | 0.5088 ± 0.0144 | Central; moderate or low-amplitude observability |

| Channel | nMAE ↓ | Raw MAE (μV) | Pearson r ↑ | Observability interpretation |
|---|---|---|---|---|
| C4 | 0.1119 ± 0.0012 | 3.610 ± 0.040 | 0.6336 ± 0.0071 | Central; moderate or low-amplitude observability |
| T4 | 0.1864 ± 0.0026 | 5.830 ± 0.082 | 0.6857 ± 0.0057 | Lateral temporal; moderate frontal observability |
| T5 | 0.1983 ± 0.0034 | 6.114 ± 0.106 | 0.5500 ± 0.0072 | Posterior/posterior-temporal; weaker phase observability |
| P3 | 0.1493 ± 0.0033 | 4.470 ± 0.098 | 0.4574 ± 0.0066 | Posterior/posterior-temporal; weaker phase observability |
| Pz | 0.1382 ± 0.0042 | 4.302 ± 0.132 | 0.3521 ± 0.0209 | Posterior/posterior-temporal; weaker phase observability |
| P4 | 0.1734 ± 0.0035 | 5.191 ± 0.104 | 0.4399 ± 0.0079 | Posterior/posterior-temporal; weaker phase observability |
| T6 | 0.2196 ± 0.0029 | 6.753 ± 0.089 | 0.5581 ± 0.0035 | Posterior/posterior-temporal; weaker phase observability |

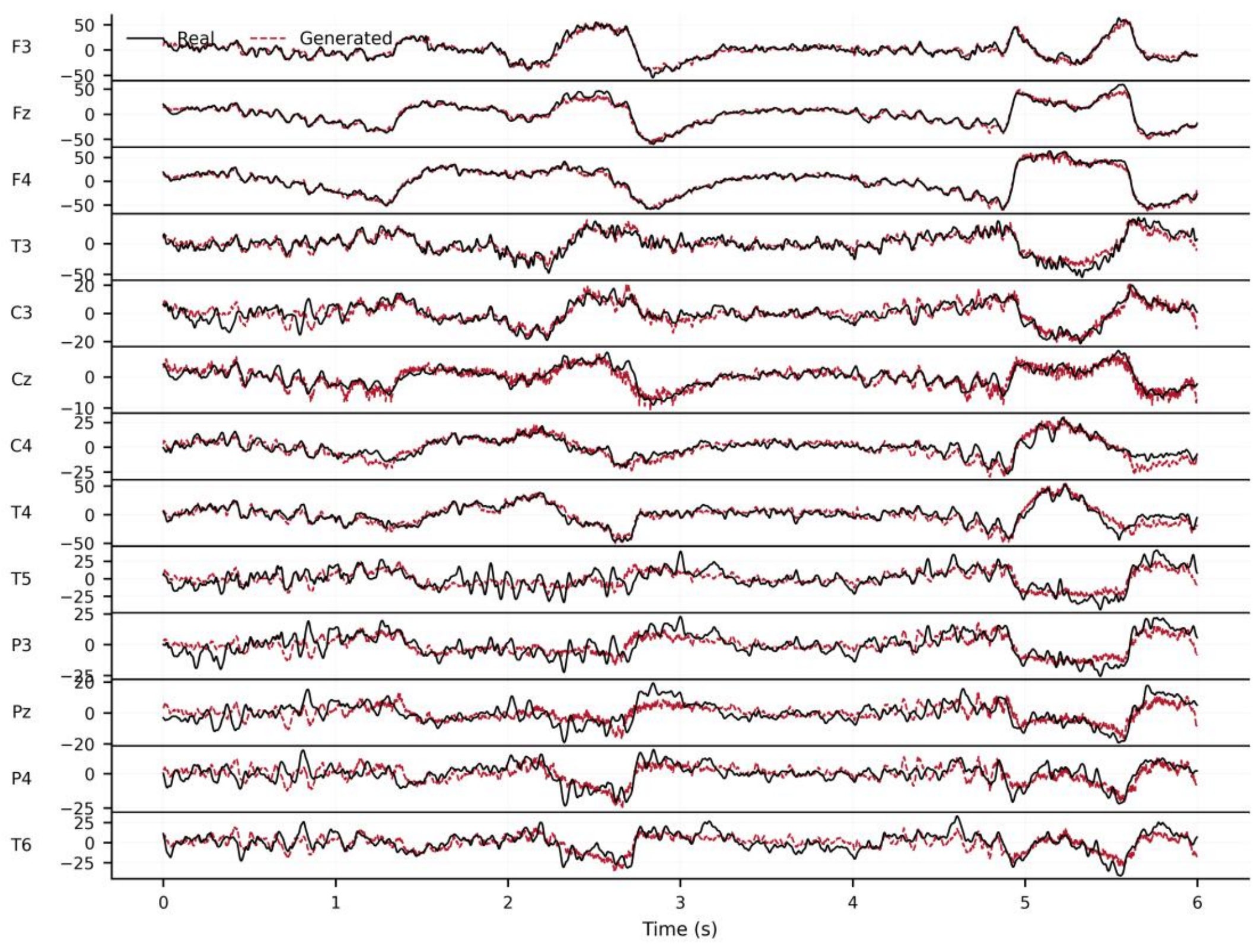

**Fig. 2 Representative overlay of real and estimated virtual EEG signals. Estimated traces follow broad morphology and local waveform patterns in many targets; posterior targets remain less phase-locked, consistent with weaker frontal observability.**

## 3.3 Spectral organization and ablation evidence

The spectral heatmap comparison in Fig. 3 and the bandpower topographic comparison in Fig. 4 provide complementary views of the main reconstruction advantage. Real targets contain channel-dependent spectral texture, high-frequency attenuation, and regional differences in energy. FAVC-Net more closely approximated these heterogeneous channel-frequency and topographic structures than interpolation baselines and several neural baselines, consistent with the large LSD and KL margins in Table 1.

The comparison is intentionally not reduced to a single averaged number. A low averaged spectral error could still hide channel homogenization or misplaced regional power. By inspecting channel-frequency heatmaps, bandpower topographies, LSD, and KL together, the analysis tests whether the model preserves both the overall spectral envelope and the heterogeneous distribution of energy across virtual channels. This is important for downstream features such as relative band power, spectral slope, and regional spectral ratios.

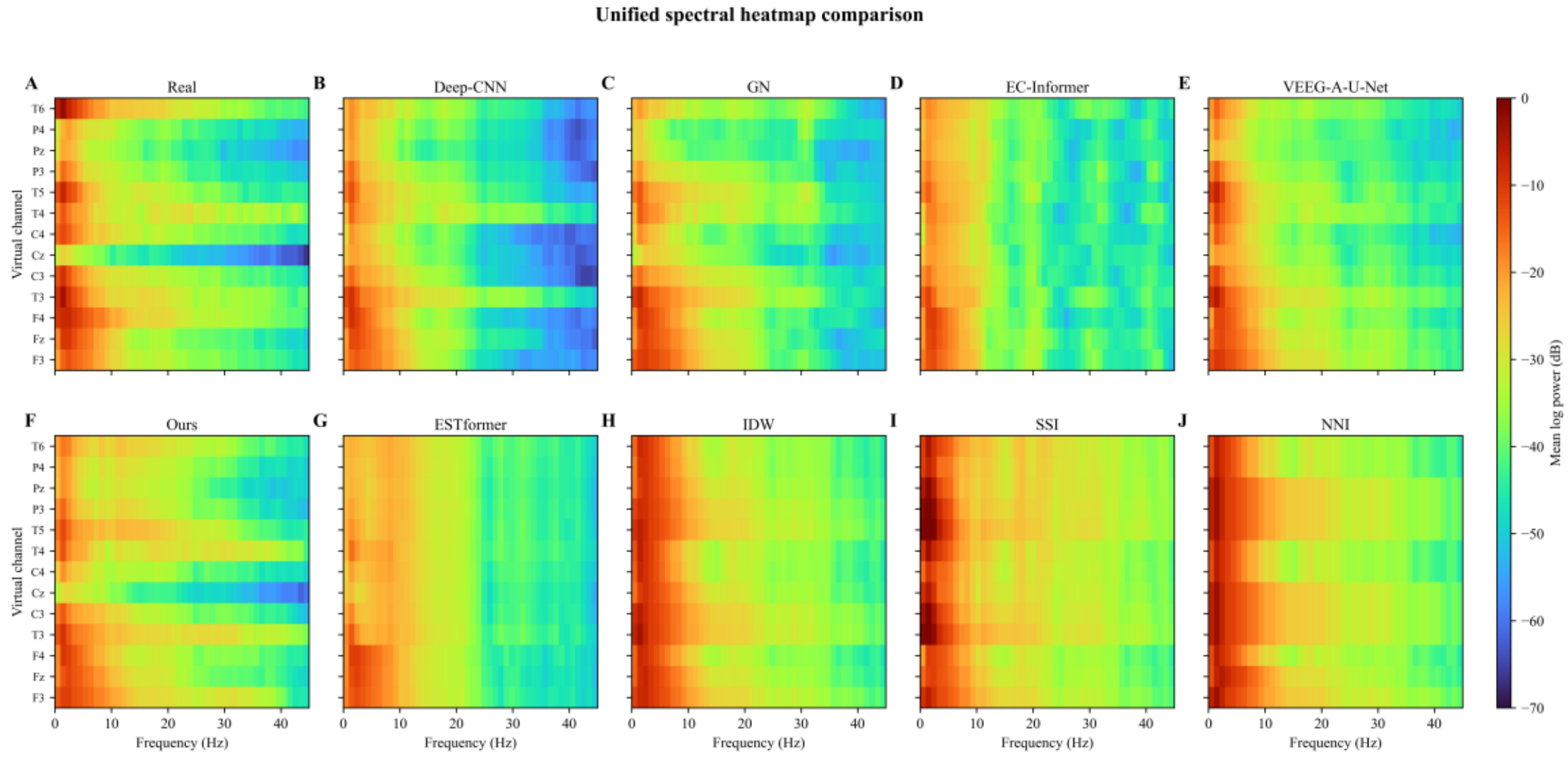

**Fig. 3 Unified spectral heatmap comparison. FAVC-Net preserves global attenuation and channel-dependent spectral texture more closely than interpolation baselines and several neural baselines.**

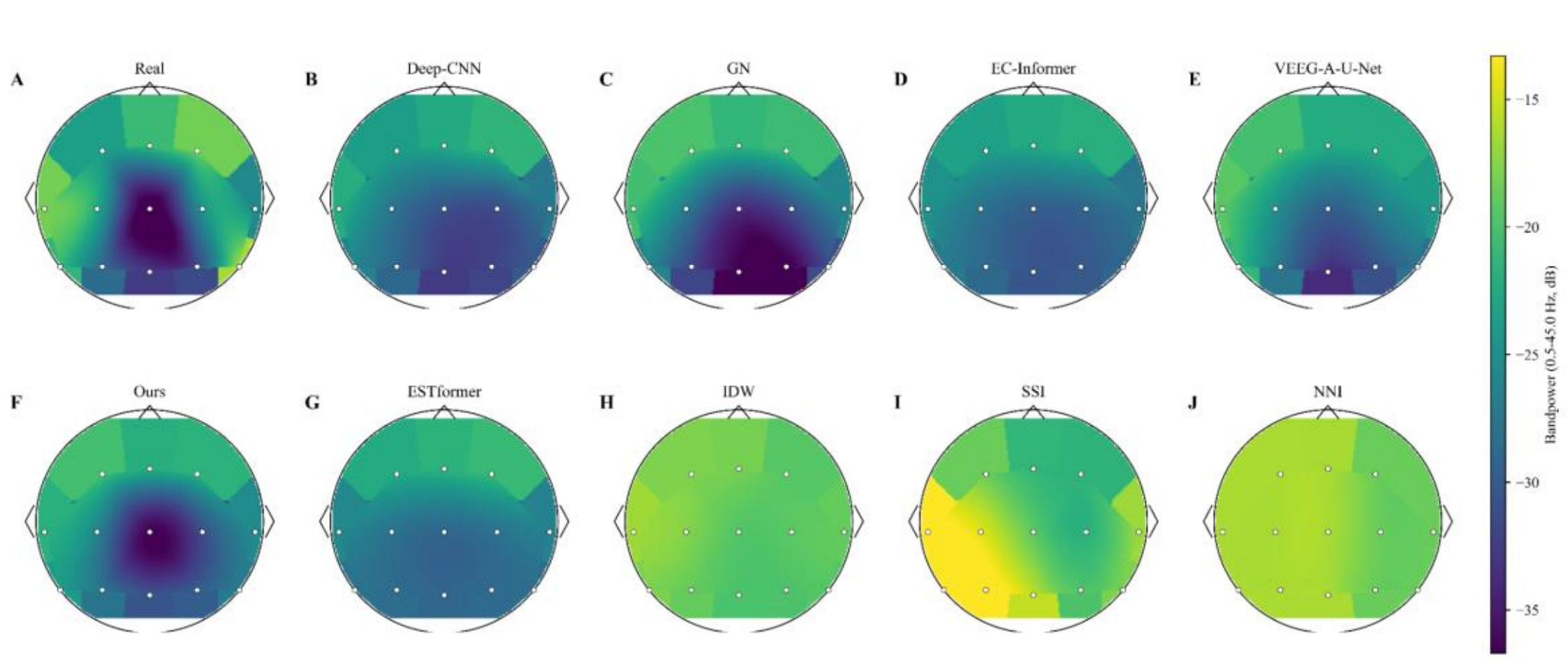

**Fig. 4 Topographic comparison of 0.5-45.0 Hz bandpower. All maps share the same color scale. FAVC-Net (Ours) best approximates the central low-bandpower shadow and the surrounding frontal/lateral-to-central energy gradient, whereas other methods either smooth the depression, shift it posteriorly or laterally, or remove it almost entirely.**

The bandpower topographies in Fig. 4 provide a spatially explicit check of this spectral advantage. The real target map contains a pronounced low-bandpower shadow over the central-to-centro-posterior scalp, surrounded by relatively higher frontal and lateral energy. This pattern is not a simple smooth interpolation from the four frontal sources: a useful conditional estimator should approximate the existence, depth, and location of the low-power basin while maintaining the surrounding energy gradient. FAVC-Net most clearly approximates this compact dark shadow at the correct midline-centered location. Although its depression is slightly rounder than the real map, its spatial position, contrast against the surrounding rim, and transition from frontal/lateral green-teal regions to the central purple basin are closest to the recorded topology among the evaluated methods.

The topographic differences among the competing methods further illustrate the distinct behavior of FAVC-Net. Deep-CNN, ESTformer, and EC-Informer mainly reproduce a smooth anterior-to-posterior bandpower gradient, with relatively broad attenuation over the posterior half of the scalp map; however, none of them clearly approximates the localized low-bandpower basin observed in the real distribution. GN and VEEG-A-U-Net better preserve this posterior low-bandpower tendency and show a more evident depression than the above methods, but their reconstructed patterns still differ from the real map in terms of spatial localization and intensity. In particular, the low-bandpower area in GN appears shifted toward the posterior boundary, whereas VEEG-A-U-Net underestimates the depth and spatial extent of the depression. The interpolation-based baselines are less consistent with the real topography: IDW and NNI largely smooth out the low-bandpower structure and produce nearly homogeneous high-power maps, while SSI introduces peripheral high-power artifacts. Overall, Fig. 4 suggests that FAVC-Net provides a more faithful approximation of both the intensity contrast and posterior spatial localization of the low-bandpower basin in the real 0.5-45.0 Hz bandpower distribution.

Ablation results (Table 3) clarify the contributions of dynamic source-state attention, signed source mixing, skip fusion, and graph refinement. Replacing signed L1 mixing with softmax attention degraded all metrics, supporting the need for negative or phase-inverted source contributions under a fixed EEG reference. Replacing the dynamic attention encoder with static source attention--implemented as learnable source-channel weights that do not depend on the current EEG segment--degraded MAE by 3.57% and Pearson r by 8.47%, indicating that source-to-target coupling should be state-dependent rather than represented by a fixed source prior. Removing the attention-consistent skip connection produced a 4.90% MAE degradation and the largest KL degradation among the newly tested variants (10.02%), showing that shallow morphology and spectral allocation are better preserved when skip information is filtered by the same attention structure used for source aggregation.

**Table 3 Structural ablation results. Degradation is reported relative to the full FAVC-Net; larger values indicate worse performance after removing the component.**

| Ablation setting | MAE degradation (%) ↑ | Pearson r degradation (%) ↑ | LSD degradation (%) ↑ | KL degradation (%) ↑ |
|---|---|---|---|---|
| Softmax attention | 10.07 | 9.17 | 3.09 | 14.11 |
| Static source attention | 3.57 | 8.47 | 0.06 | 1.53 |
| No attention-consistent skip | 4.90 | 5.62 | 0.96 | 10.02 |
| No GATv2 | 5.66 | 5.85 | 0.79 | 9.21 |
| GATv2 without spatial prior | 4.90 | 5.29 | 0.83 | 5.15 |
| No multi-scale encoder | 4.45 | 5.34 | 0.83 | 7.43 |

### 3.4 Robustness under wearable-like source-channel perturbations

Robustness results strengthen the engineering interpretation of FAVC-Net (Table 4). Across moderate EMG burst, 0.5-s dropout, ±20% gain mismatch, and moderate mixed stress, the most consistent advantages appeared in LSD, PSD KL, and CFTC. FAVC-Net kept the lowest LSD and KL and the highest CFTC among the compared methods in all four perturbed conditions. Under moderate EMG-like burst contamination, it reduced LSD by 37.2% and KL by 40.5% relative to the strongest non-FAVC comparators, reduced SCI by 8.5%, and increased CFTC by 3.0%.

The dropout and gain-mismatch conditions test different deployment risks. A 0.5-s dropout approximates transient electrode-contact instability or packet loss, whereas ±20% gain mismatch approximates impedance, calibration, or device-gain variability. In both conditions, FAVC-Net maintained large LSD and KL advantages (31.7%-44.6%) and higher CFTC (3.8%). However, SCI was slightly higher than the best non-FAVC value under dropout and gain mismatch, indicating that the robustness profile should be interpreted primarily as preservation of spectral fidelity and channel-frequency texture rather than uniformly superior anti-collapse behavior under every perturbation.

The moderate mixed-stress condition is the most deployment-facing test because it combines broadband noise, EMG bursts, short source-channel loss, and gain mismatch. Under this condition, FAVC-Net achieved LSD 0.871 ± 0.018 versus 1.241 ± 0.035 for GN, KL 0.105 ± 0.003 versus 0.153 ± 0.007, SCI 0.053 ± 0.011 versus 0.075 ± 0.009, and CFTC 0.916 ± 0.003 versus 0.913 ± 0.003. These correspond to 29.8% lower LSD, 31.5% lower KL, 28.9% lower SCI, and a 0.4% higher CFTC. Thus, under the combined perturbation that best approximates simultaneous wearable stressors, FAVC-Net preserved absolute spectral fidelity while also limiting channel-frequency collapse.

**Table 4. Robustness under wearable-like source-channel perturbations.**

| Condition | Metric | FAVC-Net | Best non-FAVC | Improvement |
|---|---|---|---|---|
| Moderate EMG burst | LSD ↓ | 0.707 ± 0.011 | VEEG-A-U-Net (1.126 ± 0.022) | 37.2% |
| Moderate EMG burst | KL ↓ | 0.083 ± 0.003 | Deep-CNN (0.140 ± 0.007) | 40.5% |
| Moderate EMG burst | SCI ↓ | 0.059 ± 0.010 | GN (0.064 ± 0.009) | 8.5% |
| Moderate EMG burst | CFTC ↑ | 0.942 ± 0.002 | GN (0.915 ± 0.003) | 3.0% |
| 0.5-s dropout | LSD ↓ | 0.665 ± 0.016 | VEEG-A-U-Net (0.974 ± 0.014) | 31.7% |
| 0.5-s dropout | KL ↓ | 0.083 ± 0.003 | Deep-CNN (0.147 ± 0.008) | 43.7% |
| 0.5-s dropout | SCI ↓ | 0.082 ± 0.012 | GN (0.074 ± 0.009) | -10.6% |
| 0.5-s dropout | CFTC ↑ | 0.949 ± 0.003 | GN (0.914 ± 0.003) | 3.8% |
| ±20% gain mismatch | LSD ↓ | 0.679 ± 0.016 | VEEG-A-U-Net (0.991 ± 0.014) | 31.5% |
| ±20% gain mismatch | KL ↓ | 0.083 ± 0.003 | Deep-CNN (0.149 ± 0.008) | 44.6% |
| ±20% gain mismatch | SCI ↓ | 0.081 ± 0.012 | GN (0.072 ± 0.009) | -13.0% |
| ±20% gain mismatch | CFTC ↑ | 0.948 ± 0.003 | GN (0.913 ± 0.004) | 3.8% |
| Moderate mixed stress | LSD ↓ | 0.871 ± 0.018 | GN (1.241 ± 0.035) | 29.8% |
| Moderate mixed stress | KL ↓ | 0.105 ± 0.003 | GN (0.153 ± 0.007) | 31.5% |
| Moderate mixed stress | SCI ↓ | 0.053 ± 0.011 | GN (0.075 ± 0.009) | 28.9% |
| Moderate mixed stress | CFTC ↑ | 0.916 ± 0.003 | GN (0.913 ± 0.003) | 0.4% |

SCI, spectral-collapse index; CFTC, channel-frequency texture correlation. Values are mean ± 95% CI. Improvement is computed relative to the listed strongest non-FAVC comparator; negative values indicate worse FAVC-Net performance for that lower-is-better metric.

### 3.5 Exploratory external representation transfer

The auxiliary external transfer results are summarized in Table 5. Adding FAVC-estimated channels to the real four-channel feature set (D1) improved segment-level AUC over the four-channel baseline (D0) for KNN, Random Forest, and RBF-SVM. D2, which used estimated virtual channels only, was also competitive, suggesting that the virtual montage forms a useful transform of the measured frontal signals. This should be interpreted as a representation-compatibility and finite-sample modeling effect, not as proof that the virtual channels contain independent measurement information beyond the source channels. Importantly, the negative controls D3 and D4 did not consistently match D1, indicating that the gains are not explained only by adding more columns or by arbitrary pseudo-channel spectra.

The main value of this experiment is not the absolute performance of an external classifier. Rather, it connects reconstruction quality with a realistic engineering use case: a low-channel wearable recording can be mapped into a virtual montage whose spectral-topographic features are more compatible with multichannel analysis pipelines under device, montage, sampling-rate, and cohort mismatch. Because the cohort was small and segment-level samples within subject are dependent, these findings are reported as exploratory representation-transfer evidence rather than clinical validation.

**Table 5 Exploratory external representation-transfer check. D0 uses real four-channel features; D1 adds FAVC-estimated virtual channels; D2 uses estimated virtual channels only; D3 and D4 are dimensionality and pseudo-spectrum controls.**

| Classifier | D0 AUC | D1 AUC | D2 AUC | D3 AUC | D4 AUC | D1-D0 AUC | D1 balanced accuracy |
|---|---|---|---|---|---|---|---|
| KNN | 0.6928 | 0.7528 | 0.7123 | 0.7165 | 0.7209 | +0.0600 | 0.7142 |
| Random Forest | 0.6710 | 0.7369 | 0.7340 | 0.6814 | 0.6770 | +0.0659 | 0.6830 |
| RBF-SVM | 0.6472 | 0.7105 | 0.6975 | 0.6697 | 0.6745 | +0.0633 | 0.6638 |

Results are segment-level summaries from subject-held-out folds and are reported as exploratory representation-transfer evidence rather than clinical validation.

## 4. Discussion

This study reframes four-frontal virtual EEG channel estimation as prior-guided conditional inference of unobserved scalp-potential samples and as a dual-domain reconstruction problem. The output is a deterministic posterior predictive representation learned from multichannel EEG data, not an additional physical measurement. The most important empirical pattern is the separation between modest waveform gains and substantial spectral gains. FAVC-Net does not claim to recover all posterior and parietal phases from four frontal sensors. Instead, it preserves waveform morphology where observable and substantially improves spectral-topographic reliability, which is the form of information most relevant to many wearable EEG features.

The architectural results clarify why the gain occurs. A shared multi-scale encoder extracts compact source representations; source-state embeddings make the source-to-target mapping adaptive to the current window; target-private attention layers give each virtual electrode its own response rule; signed L1 mixing allows negative dependencies; and GATv2 refinement lets neighboring virtual targets exchange information about source-block selection without directly smoothing voltages. The ablation results show complementary roles for these components: static source attention weakens phase-sensitive coupling because it ignores the current EEG segment, attention-consistent skip removal harms waveform fidelity and spectral allocation, softmax attention suppresses useful signed dependencies, and GATv2 with its spatial prior provides smaller but consistent stabilization. The topographic bandpower analysis supports the same interpretation by showing that the full model approximates the central low-power shadow of the real montage rather than producing an over-smoothed or displaced energy map. This hierarchy is important because it explains the model as an interpretable conditional mapping from measured source states to target-specific source-block mixtures rather than as a large generic neural generator.

The same structure provides a cautious form of extensibility. Since most computation is shared and each target is mainly specified by a compact attention generator and spatial metadata, the model is not tied as strongly to a monolithic fixed-output decoder as many multi-channel generators. This property is useful for scenarios in which different studies or wearable deployments require different virtual-channel subsets. However, the graph refinement stage couples target attention patterns, and new target locations would still require appropriate coordinates and training or fine-tuning data. The present evidence should therefore be read as architectural modularity within a validated 13-channel target set, not as zero-shot recovery of arbitrary electrodes.

The information-theoretic interpretation also constrains the downstream claim. A deterministic virtual montage cannot increase the Shannon information contained in a single measured source segment after the generator is fixed. Downstream improvements can nevertheless occur because real classifiers are finite-sample, architecture-constrained, and sensitive to montage mismatch. FAVC-Net may help by embedding learned multichannel priors, making spatial-spectral features explicit, and presenting low-channel recordings in a form more compatible with multichannel feature extractors. This interpretation is narrower but more defensible than describing virtual channels as new independent observations.

The robustness analysis is central to biomedical-engineering relevance. A model trained and tested only under clean public-dataset conditions could be viewed as another offline AI benchmark. Here, perturbations were applied only to the measured source channels, while the target channels remained clean, matching the practical question of whether a wearable system can infer a stable virtual montage from imperfect inputs. The strongest evidence lies in absolute noisy-condition LSD, PSD KL, and CFTC, with SCI providing condition-specific anti-collapse evidence under EMG-like and mixed stress. These endpoints show that FAVC-Net is less prone to spectral distortion and better preserves channel-frequency texture under corrupted-source conditions, while dropout and gain-mismatch SCI results indicate that spectral-collapse behavior should be interpreted condition by condition rather than as a uniform advantage.

The channel-wise results define the method's limitations. Frontal targets were highly observable, whereas posterior and parietal targets had weaker phase-sensitive correlation. This should not be hidden because it is a fundamental limitation of sparse sensing. In practice, posterior and parietal virtual channels should be interpreted primarily as frequency-calibrated representations derived from frontal measurements rather than as new independent physiological observations. This observability-aware interpretation reduces the risk of overclaiming and makes the method more suitable for engineering deployment.

The external transfer check is consistent with this interpretation. D1 improvements over D0 suggest that the fixed FAVC montage can provide a more useful representation for finite-sample downstream models than the raw four-channel feature set alone, while negative controls reduce the likelihood that gains arise only from feature-dimensionality expansion. However, the external cohort was small, labels were not used to train the generator, and segment-level samples were subject-dependent. Therefore, the downstream experiment should be read as a bridge between reconstruction and potential application, not as clinical validation or evidence of independent information creation.

This interpretation also defines inappropriate uses. Virtual channels should not be treated as directly recorded evidence for new regional neural activity, source localization, scalp connectivity, or causal neurophysiological claims unless supported by physical recordings and additional validation. Learned priors, volume conduction, shared artifacts, and reference effects can all induce apparent cross-channel structure. The generated montage is most defensible as a representation-completion and model-adaptation interface for low-channel wearable EEG, with reliability depending on target observability, frequency band, artifact state, and distribution match.

Several limitations remain. The primary reconstruction experiment used a public task-specific dataset with young adult participants, and broader validation is needed across age ranges, tasks, devices, references, and artifact profiles. The robustness tests were synthetic but wearable-informed; real ambulatory artifact corpora would provide stronger deployment evidence. The current model is deterministic and does not quantify posterior uncertainty, so it cannot directly indicate when a target channel is weakly inferable from the observed sources. The method also requires basic quality control of the measured frontal channels and should not be used to infer unobserved physiology when source channels are saturated or missing for long periods. Future work should evaluate uncertainty estimation, online quality gating, prospective wearable recordings, montage-variable target sets, and downstream models trained explicitly to account for virtual-channel reliability.

## 5. Conclusion

FAVC-Net provides a compact prior-guided, frequency-calibrated framework for estimating 13 virtual EEG channel representations from four frontal electrodes. Its main contribution is a robust spectral-topographic reconstruction advantage and a montage-compatible representation interface, not a claim of perfect waveform recovery or independent physical measurement. Target-conditioned signed source-block mixing, GATv2 attention refinement, and weak PSD calibration jointly improve spectral fidelity, preserve channel-frequency texture, and maintain useful waveform information where observable. The corrupted-source robustness results further support the method as a practical representation-completion module for low-channel wearable EEG systems, especially when absolute spectral fidelity is the primary endpoint.

## Declarations

**Funding:** No funding was received for this study.

**Ethics approval and consent to participate:** The primary reconstruction analysis used the public PRED+CT EEG dataset retrospectively. The exploratory external transfer check used previously collected four-channel EEG recordings from the low-cost wireless EEG measurement system described by Yu and Guo [34]. No new participant recruitment, intervention, or prospective data collection was performed for the present methodological study.

Data availability: The primary PRED+CT EEG data used for reconstruction are publicly available through the Patient Repository of EEG Data + Computational Tools (PRED+CT; https://predict.cs.unm.edu/), a public EEG repository described at DOI: 10.3389/fninf.2017.00067. The PRED+CT study analyzed here is reported by Cavanagh et al. [4] (DOI: 10.1162/cpsy_a_00024). The exploratory external four-channel EEG data are from the low-cost wireless EEG measurement system described by Yu and Guo [34]; access may be provided upon reasonable request to the corresponding author, subject to institutional and ethical restrictions.

**Code availability:** The implementation code and trained checkpoints are available upon request from the corresponding author. Please direct inquiries to E-mail: 3098975303@qq.com (the email most frequently checked by the author) to ensure a timely response.